\documentclass[12pt]{article}
\usepackage{float}
\usepackage{graphicx}
\usepackage{subcaption}
\usepackage{amsmath}
\usepackage{amsopn}
\title{Principal Component Analysis\\
Stochastic Optimization}

\author{Jian Vora, IIT Bombay}
\date{\vspace{-5ex}}
\begin{document}

\maketitle
\section{Abstract}
Principal Component Analysis (PCA) is a novel way of of dimensionality reduction. This problem essentially boils down to finding the top k eigen vectors of the data covariance matrix. A considerable amount of literature is found on algorithms meant to do so such as an online method be Warmuth and Kuzmin, Matrix Stochastic Gradient by Arora, Oja's method and many others. In this paper we see some of these stochastic approaches to the PCA optimization problem and comment on their convergence and runtime to obtain an $\epsilon$ sub-optimal  solution. We revisit convex relaxation based methods for
stochastic optimization of principal component
analysis (PCA). While methods that directly solve
the nonconvex problem have been shown to be
optimal in terms of statistical and computational
efficiency, the methods based on convex relaxation have been shown to enjoy comparable, or
even superior, empirical performance – this motivates the need for a deeper formal understanding
of the latter.
\section{Introduction}
Principal Component analysis (PCA) is a tool which is used in data analysis, machine learning and many other applications. The increasing amount of data at our disposal has increased the computational costs in drawing useful conclusions from it, for instance training a model. PCA is used to lower the dimensionality of this big data available so as to both reduce the cost and improve our visualization for the data.\\\\
Mathematically, given n vectors in R\textsuperscript{d}, PCA deals with finding the k-dimensional subspace which captures the maximum variance of the data. Here the distribution D from which the samples are drawn is unknown. Given the n x d data matrix X, it is well known that the spanning vectors of the subspace are the leading k eigen vectors of X (or the covariance matrix if the data is centered). This problem can be posed as finding the singluar value decomposition (SVD) of n x d data matrix. Hence most of the work on PCA deals which effective and compuataionally cheaper ways of evaluating the SVD of X. Summing up, PCA defines a new orthogonal coordinate system that optimally describes variance in a single dataset.\\\\
The straightforward approach is “Sample Average Approximation” (SAA), where we collect a sample of data points, and then optimize an empirical version of the objective on the sample using standard deterministic techniques. This is essentially an Empirical Risk Minimization (ERM) approach. However, it is the computational costly. For instance, considering a problem dealing with data sets of the size of the population of a country, then Batch Gradient Descent will iterate through all the records for just a single parameter update. We calculate the second-moment matrix M =$\sum_{t=1}^{T} x\textsubscript{t}x\textsubscript{t}\textsuperscript{T}/T $ where T is the number of samples we use. Subspace spanning vectors are the leading k eigen vectors of M. Eigen decomposition of M takes $\mathcal{O}(d\textsuperscript{3})$ time while calculating M itself takes $\mathcal{O}(Td\textsuperscript{2})$ which is pretty expensive when we have huge data sets.\\\\
In this paper, we look at and alternative stochastic approximation approaches to the PCA optimization problem. Here, we consider each individual vector iteratively for performing the update. Next, we shall analyze few stochastic approximation algorithms and look at the complexities.

\section{Problem Statement}
PCA is essentially finding a k-dimensional subspace which captures the maximum variance of the data.  It can be posed as an optimization problem. If x $\in$ R\textsuperscript{d} and U $\in$ R\textsuperscript{d $\times$ k}, where the columns of U contain the orthonormal spanning vectors of the desired subspace.
\begin{center}
   \textbf{ maximise : E\textsubscript{x}[trace(U\textsuperscript{T}xx\textsuperscript{T}U)]\\
    subject to : U\textsuperscript{T}U $\preceq$ I}
    
\end{center}
The above expresion of expectation comes from taking singluar value decomposition of x. Here, expectation is taken with repected to the unknown distribution D. In the SAA approach described above, expectation is replaced by sample mean.The regime we consider in this paper is where we have an essentially unlimited supply of training
examples, and would like to obtain an $\epsilon$-suboptimal solution
in the least possible runtime.\\
Now we discuss some novel algorithms for PCA in the stochastic setting. Section 3.1 discusses Incremental SVD method for low-rank approximations. Section 3.2 discusses power method where we discuss efficient ways to solve SGD by trying to avoid renormalization at every step. Last we discuss Matrix Stochastic Gradient and a refined form of it in terms of capped MSG where the rank of the iterates isn't allowed to blow up.

\section{Methods}
\subsection{Incremental Singluar Value Decompositon}
SVD is a matrix factorization technique commonly used for producing low-rank approximations. Given a m x n matrix A, then it can be decomposed as follows:\\
\begin{center}
   \textbf{ A = U$\sum$V\textsuperscript{T}}
\end{center}
Here U,$\sum$ and V are of dimensions m x m, m x n and n x n respectively. If the rank of matrix A was r then we can drop the zero entries rows/columns of $\sum$ and the new dimensions of the matrices as m x r, r x r and r x n respectively. $\sum$ is a diagonal matrix with entries arranged decreasing as we move down the diagonal. If we need a rank k approximation of A then we need to take the first k x k submatrix of $\sum$, first k columns of U and first k rows of V and multiply them in the same order to get the \textbf{low-rank approximation} of A. This new low rank matrix obtained, say A\textsubscript{k}, is the best approximation to A. Specifically, A\textsubscript{k} minimizes the Frobenius Norm with respect to A.\\\\
Given the background of SVD, what we propose to do in incremental SVD is to take each data point at a time and update the matrix of eigen vectors. Suppose we have reached a l-rank approximation and need a k-rank then given a new data point, we follow the following algorithm:\\
Inputs:U,S of the existing matrix, desired rank k of final matrix, data x\\
Outputs:U,S of the updated matrix\\
Note : For notation Q and U\textsuperscript{'} and S\textsuperscript{'} are defined in the paper 1 of the references.\\
\textbf{Algorithm:}\\
1. Project data on U, find it's orthogonal complement and normalise it\\
2. Formulate the matrix Q and perform an eigen decomposition on it\\
3. Find the new U and S as described earlier\\
4. Truncate by taking the top rank number of singular values and their corresponding vectors\\\\
Such an incremental approach has a space complexity of $\mathcal{O}$(kd) and a time complexity of $\mathcal{O}$(Tk\textsuperscript{2}d). There are cases where this method fails to converge especially when the input data is orthogonal. One advantage of it is a better performance in terms of sub-optimality versus number of iterations. Another advantage of this approach is that it is independent of the step  size or learning rate parameter.
\subsection{Stochastic Power Method}
Assuming D is unknown
but unchanging, the goal is to update U directly as
samples x\textsubscript{i} arrive sequentially and independently
The update proposed in this algorithm is :\\
\begin{center}
    
\textbf{U\textsuperscript{(t+1)}=$\mathcal{P}$\textsubscript{orth}(U\textsuperscript{(t)} + $\eta$\textsubscript{t}x\textsubscript{t}x\textsubscript{t}\textsuperscript{T}U\textsuperscript{(t)})}
\end{center}
Here $\eta$\textsubscript{t} is the learning rate and $\mathcal{P}$\textsubscript{orth} projects its argument onto the set of
rank-k matrices with respect to the spectral norm. This can
be accomplished by setting all but the k largest eigenvalues
to zero. Convergence is guaranteed but the rate of convergence is unknown. The complexity of this algorithm is as follows: Projection can be done in $\mathcal{O}$(k\textsuperscript{2}d) time using Gram-Schmidt or QR factorization. Time-complexity of this algorithm is $\mathcal{O}$(Tkd) and a space complexity of $\mathcal{O}$(kd). This is an improvement of a factor of k/d as compared to the conventional batch algorithms. There has been a lot of literature of the same SGD update but they vary in they way they renormalize. For instance, Oja and Karhunen perform orthonormalization after every iteration, while the
popular generalized Hebbian algorithm, which was later
generalized to the kernel PCA setting
performs a partial renormalization.
\subsection{Matrix Stochastic Gradient}
Here we consider a re-parameterization of the PCA objective. Instead of looking at the matrix of basis vectors U,we now consider the projection matrix M = UU\textsuperscript{T}. M is thus a d x d square matrix. The modified optimization problem is specified as follows:\\
\begin{center}
    \textbf{ maximise : E\textsubscript{x}[x\textsuperscript{T}Mx]\\
    subject to : rank(M)=k ; $\lambda$\textsubscript{i}(M)}$\in$ 0,1
\end{center}
The above problem is not convex because of it's constraints. We can relax the constraint by taking convex hull of the problem. Thus the new constraint becomes trace(M) = k and 0$\preceq$M$\preceq$I. Warmuth and Kuzmin (2008) showed that and solution of the relaxation can be expressed as a convex combination of at most d feasible solutions of the original problem.\\
The update use for MSG is pretty similar to the Stochastic Power Method and is describes as below:\\
\begin{center}
\textbf{M\textsuperscript{(t+1)}=$\mathcal{P}$\textsubscript{orth}(M\textsuperscript{(t)} + $\eta$\textsubscript{t}x\textsubscript{t}x\textsubscript{t}\textsuperscript{T})}

\end{center}
The algorithm can be described as below:\\
1. Choose the learning rate  $\eta$, number of iterates T and initial value of M.\\
2. Iterate over all the data points individually taking one at a time.\\
3. Find the average over all the iterates and extract a rank-k solution from it.\\\\
A direct application of the algorithm gives a run-time of $\mathcal{O}$(Td\textsuperscript{2}) and a space-complexity of $\mathcal{O}$(d\textsuperscript{2}). There exists a novel algorithm where we maintain eigen decomposition of each iterate and update according to the new data point streamed.\\\\
If $\sigma$\textsubscript{i}\textsuperscript{'} are the eigen values of the t-th iterate then the updated matrix after projection on the feasible region is given by a unique matrix whose eigen values are given by the following rule:\\
\begin{center}
   \textbf{ $\sigma$\textsubscript{i} = max(0, min(1, $\sigma$\textsubscript{i}\textsuperscript{'}+ S)}
    
\end{center}
\textbf{Where S $\in$ R such that $\Sigma$ $\sigma$\textsubscript{i}=k.\\}
The eigenvectors of the new projected matrix remain the same as the previous one. Thus the eigen values of the new matrix can be obtained by shifting and clipping the values between 0 and 1. This requires a run time of $\mathcal{O}$(Tdk\textsuperscript{2}) and has a space complexity of $\mathcal{O}$(dk) where k is the rank of the iterate which we are going to project. This is a considerable improvement as compared to the conventional projection especially when the rank of iterates is small. This motivates to keep a check on the rank of the iterates and leads to the next method of capped MSG.
\section{Capped MSG}
The optimization objective an constraints are the same as that of MSG except that we additionally want to put of 'cap' or an ubber bound to the rank of the iterates. Let this bound be K. When K=k then this is essentially the incremental approach. If we set K greater than k then the performance is almost like that of incremental but has a better convergence than the incremental approach due to a larger space to search for. Generally, we set K=k+1 and takes the best of incremental and MSG approach to give a convergent solution in lesser runtime.
\section{Experimental Results}
The above described methods were tested against various data sets for their rank of iterates and suboptimality as a function of runtime and the number of iterations. Apart from the above mentioned algorithms, results are also compared against l1, l2 and l21 MSG as mentioned in paper 13 of the references. We consider data drawn from orthogonal distribution with N = 10,000 and D = 32.  As a process of pre-processing we center the data by subtracting the mean and dividing by the standard deviation to normalise the data. We want to represent the data in only  4 dimensions hence k = 4. Data was divided into 3 parts - training, tuning and testing. The plots obtained for all the methods are shown below for this synthetic data set:[legend for the first plot is applicable for all the other plots]\\
\begin{figure}[H]
    \centering
    \includegraphics[width=13cm]{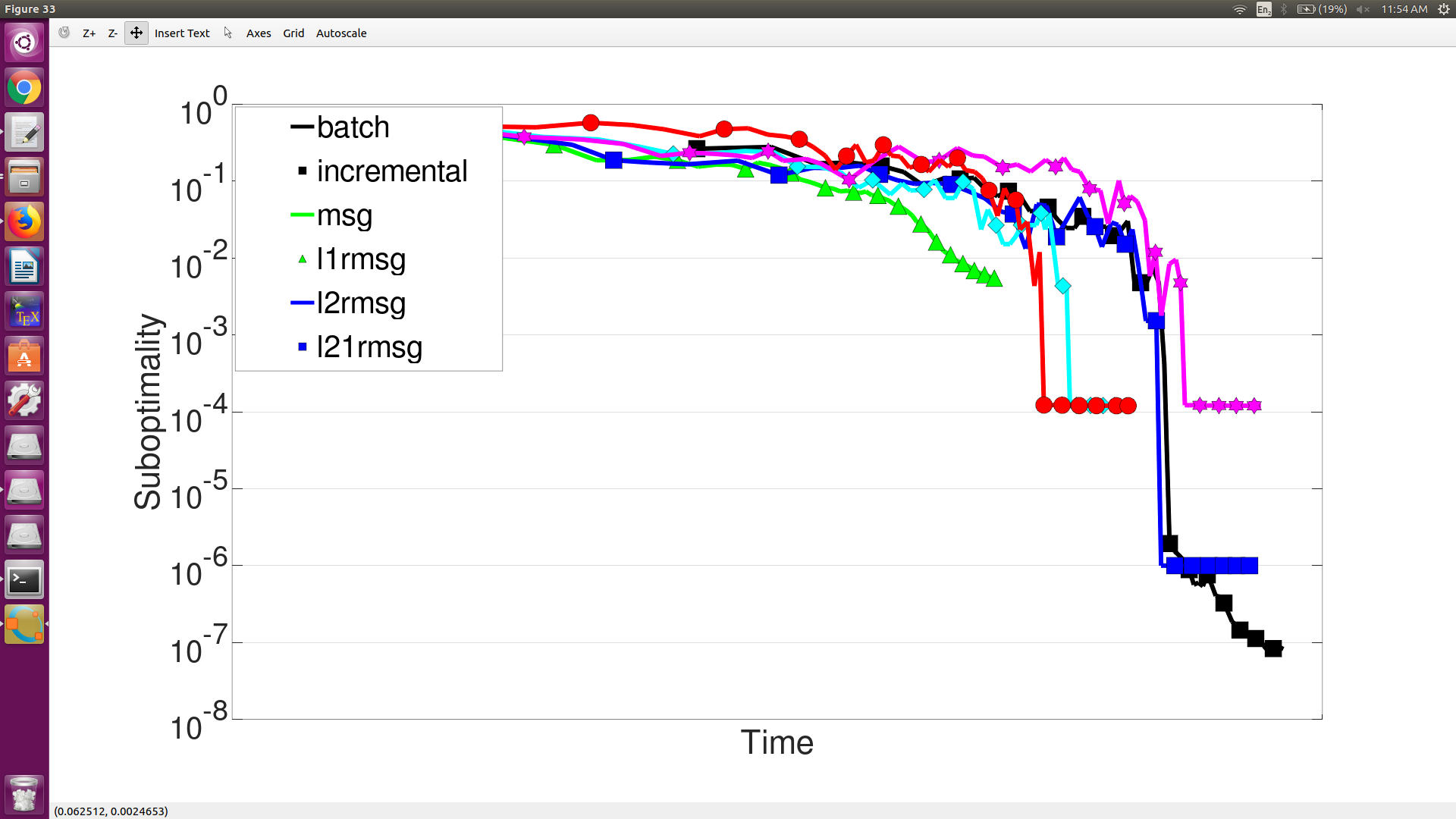}
    \caption{Suboptimality of the objective with respect to the runtime for all the various methods discussed}
    \label{fig:my_label}
\end{figure}
\begin{figure}[H]
    \centering
    \includegraphics[width=13cm]{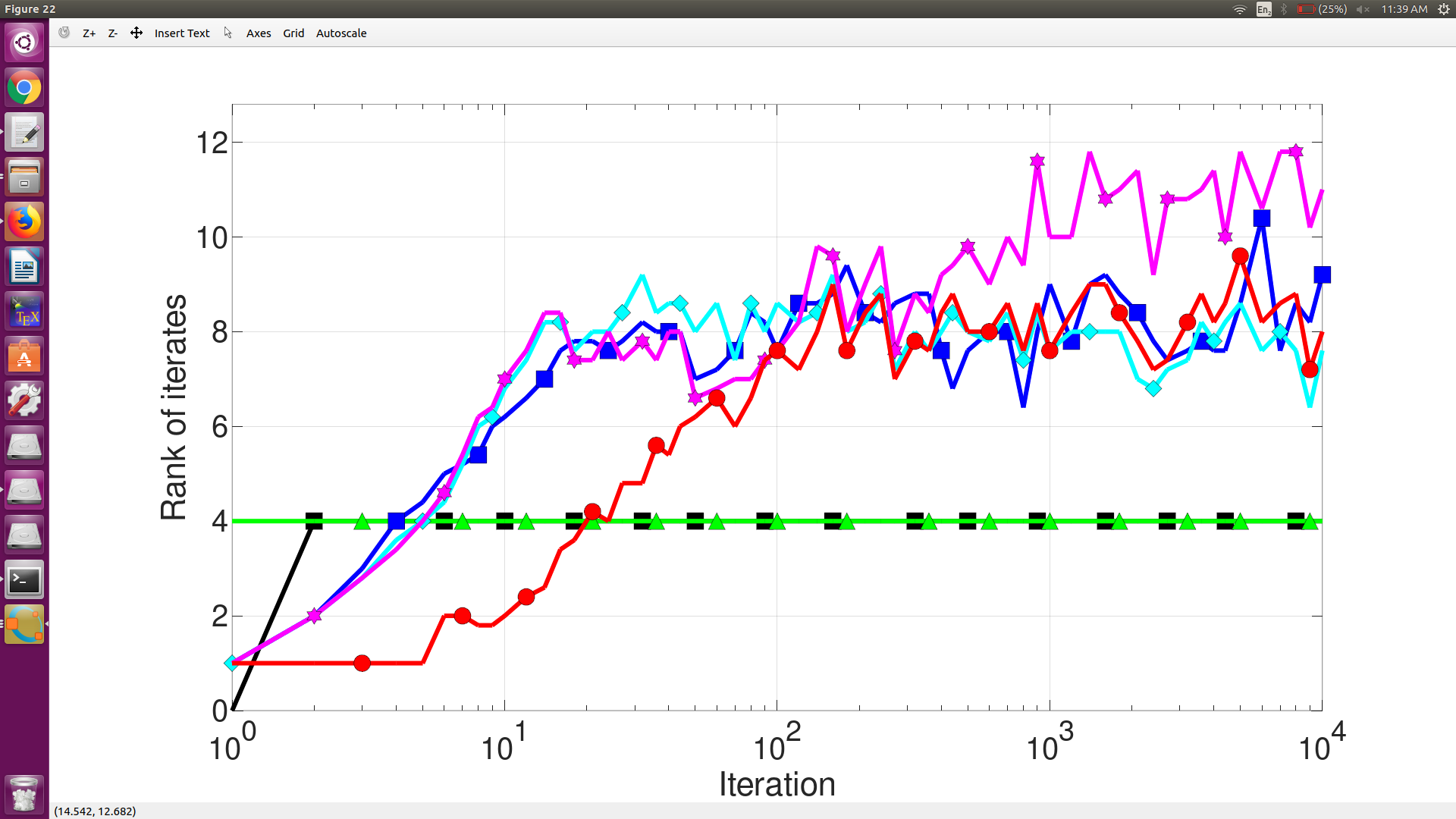}
    \caption{Rank of the iterates with respect to the number of iterations for all the various methods discussed}
    \label{fig:my_label}
\end{figure}
\begin{figure}[H]
    \centering
    \includegraphics[width=13cm]{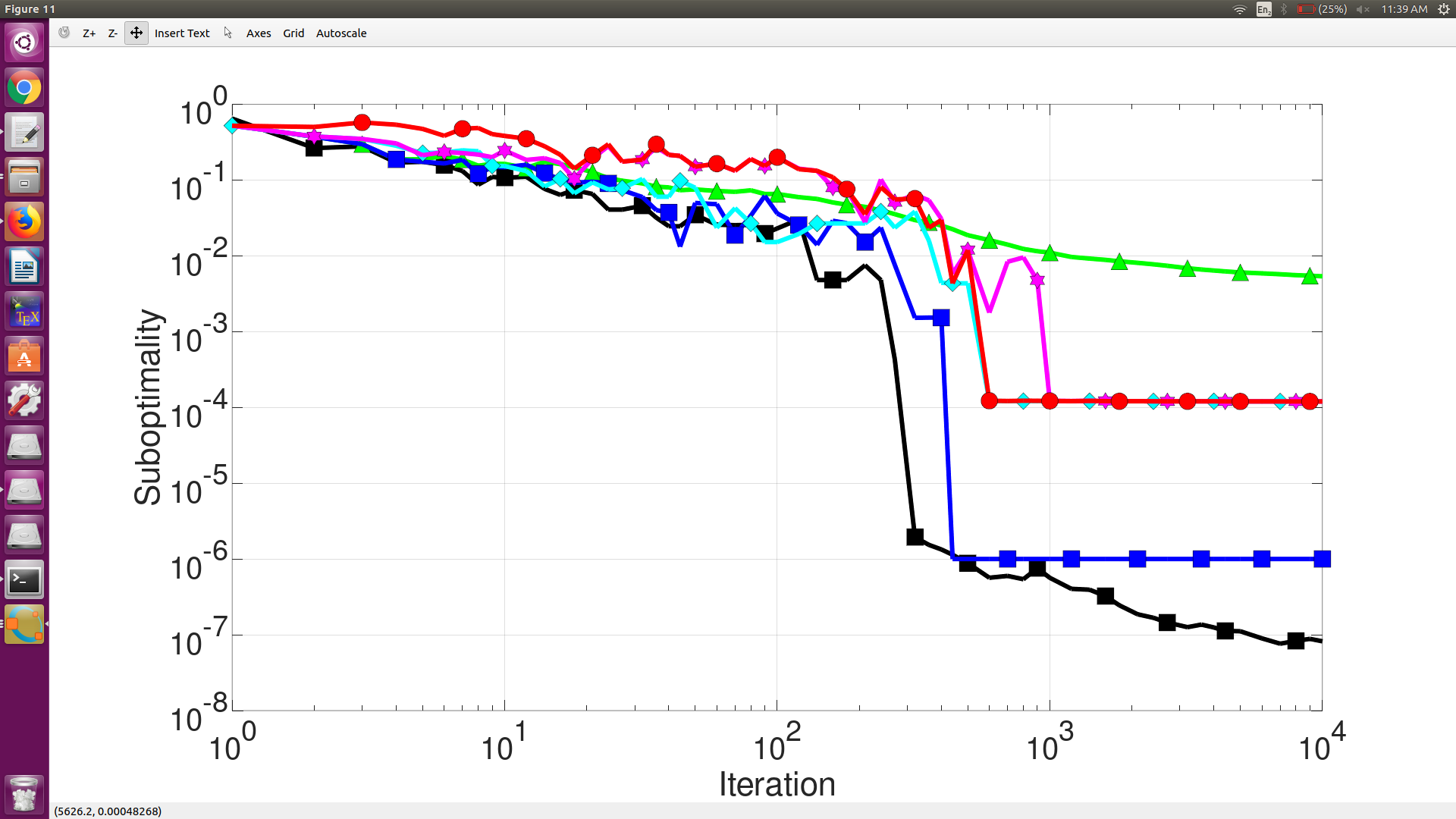}
    \caption{Suboptimality of the objective with respect to the number of iterations for all the various methods discussed}
    \label{fig:my_label}
\end{figure}
\section{Conclusion}
\textbf{Iteration Complexity:}\\
We observe that l1 and l21 rmsg achieve a better suboptimality with the number of iterations and performs much better as compared to the standard msg algorithm. Oja's algorithm, although having a faster convergence rate does not perform that well in practice as validated by previous studies.\\\\
\textbf{Rank of Iterates:}\\
Rank of msg and l1 rmsg stays a constant at 4 as they were defined to maintain a constant rank.l1 and l21 rmsg maintain a better bound of the iterate ranks as compared to the other discussed algorithms.\\\\
\textbf{Overall Runtime:}\\
Batch algorithm  achieves the best suboptimality given the number of iterations but the run-time is much more as compared to their stochastic counterparts. Matrix Stochastic Gradient has the least runtime but again at the cost of a greater deviation from the exact solution.\\\\
Thus in this paper we looked and reviewed certain algorithms for finding the k-dimensional subspace which maximises variance of an uncentered data. We tried to find solutions which even though not exact but would be computationally efficient to apply in a practical scenario. We looked at various convex relaxations of the PCA objective optimization problem which leads to various methods as discussed. We did achieve solutions which are linear in terms of d, dimension of the input data set. Further, we analysed the theoretical results obtained and compared it by testing those against a synthetic orthogonal data set.
\section{References}
1. Arora, Raman, Cotter, Andrew, Livescu, Karen, and Srebro, Nathan. Stochastic optimization for
PCA and PLS. In 50th Annual Allerton Conference on Communication, Control, and Computing,
2012.\\\\
2. E. Oja and J. Karhunen, “On stochastic approximation of the
eigenvectors and eigenvalues of the expectation of a random
matrix,” Journal of Mathematical Analysis and Applications,
vol. 106, pp. 69–84, 1985.\\\\
3. Brand, Matthew. Incremental singular value decomposition of uncertain data with missing values.
In ECCV, 2002.\\\\
4. Warmuth, Manfred K. and Kuzmin, Dima. Randomized online PCA algorithms with regret bounds
that are logarithmic in the dimension. Journal of Machine Learning Research (JMLR), 9:2287–
2320, 2008.\\\\
5. K. I. Kim, M. O. Franz, and B. Sch¨olkopf, “Iterative kernel
principal component analysis for image modeling,” IEEE
Trans. PAMI, vol. 27, no. 9, pp. 1351–1366, 2005.\\\\
6. Oja, Erkki and Karhunen, Juha. On stochastic approximation of the eigenvectors and eigenvalues
of the expectation of a random matrix. Journal of Mathematical Analysis and Applications, 106:
69–84, 1985.\\\\
7. Stochastic Optimization of PCA with Capped MSG.Raman Arora, Andy Cotter, Nati Srebro NIPS 2013\\\\
8. Warmuth, Manfred K and Kuzmin, Dima. Randomized
online PCA algorithms with regret bounds that are logarithmic in the dimension. Journal of Machine Learning
Research, 9(10), 2008.\\\\
9. Balsubramani, Akshay, Dasgupta, Sanjoy, and Freund, Yoav.
The fast convergence of incremental PCA. In Advances in
Neural Information Processing Systems, pp. 3174–3182,
2013.\\\\
10. Shamir, Ohad. Fast stochastic algorithms for SVD and PCA:
Convergence properties and convexity. In International
Conference on Machine Learning, pp. 248–256, 2016.\\\\
11. Arora, Raman, Mianjy, Poorya, and Marinov, Teodor.
Stochastic optimization for multiview representation
learning using partial least squares. In International Conference on Machine Learning, pp. 1786–1794, 2016.\\\\
12. Mitliagkas, Ioannis, Caramanis, Constantine, and Jain, Prateek. Memory limited, streaming PCA. In Advances in
Neural Information Processing Systems, pp. 2886–2894,
2013.\\\\
13. Poorya Mianjy and Raman Arora. Stochastic PCA with l1 and l2 regularization. In Proceedings of the 35th International Conference on Machine Learning (ICML), 2018
\end{document}